\documentclass{article}
    \PassOptionsToPackage{numbers, compress}{natbib}

    \usepackage[preprint]{neurips_2025}

\usepackage[utf8]{inputenc} 
\usepackage[T1]{fontenc}    
\usepackage{hyperref}       
\usepackage{url}            
\usepackage{booktabs}       
\usepackage{amsfonts}       
\usepackage{nicefrac}       
\usepackage{microtype}      
\usepackage{xcolor}         
\usepackage{graphicx}

\usepackage{amssymb}
\usepackage{amsmath}
\usepackage{amsthm}
\newtheorem{definition}{Definition}
\newtheorem{theorem}{Theorem}
\newtheorem{remark}{Remark}
\newtheorem{lemma}{Lemma}

\usepackage{multirow}

\usepackage{algorithm}
\usepackage{algorithmic}

\def\eg{\emph{e.g.}}
\def\ie{\emph{i.e.}}
\usepackage{bm}
\usepackage[capitalize,noabbrev]{cleveref}

\usepackage{bbding}
\usepackage{colortbl}
\usepackage{multirow}

\title{DiffoRA: Enabling Parameter-Efficient Fine-Tuning via Differential Module Selection}

\author{%
  Tangyu Jiang\\
	Tsinghua University\\
	\texttt{jiangtangyu@sz.tsinghua.edu.cn} \\
	\And
	Haodi Wang \\
	City University of Hong Kong\\
    Lab for AI-Powered FinTech, HK\\
	\texttt{haodiwang@hkaift.com} \\
	\AND
	Chun Yuan\thanks{Corresponding author.} \\
	Tsinghua University\\
	\texttt{yuanc@sz.tsinghua.edu.cn} \\
}

\begin{document}

\maketitle

\begin{abstract}
The Parameter-Efficient Fine-Tuning (PEFT) methods have been extensively researched for large language models in downstream tasks. 
Among all the existing approaches, the Low-Rank Adaptation (LoRA) has gained popularity for its streamlined design by incorporating low-rank matrices into existing pre-trained models.
Though effective, LoRA, as well as its adaptive optimizations, either allocate the same matrix to all the modules or adjust the interior rank of the components based on importance scoring indicators.
%
%
In this paper, we argue that not all the modules in LLMs are suitable and necessary to be fine-tuned.
Enlightened by this insight, we propose a new PEFT scheme called DiffoRA, which enables adaptive adoption of the low-rank decomposition matrices.
At the core of DiffoRA lies a Differential Adaptation Matrix (DAM) to determine which module is the most suitable and essential for fine-tuning. 
We theoretically explain how the designed matrix impacts the convergence rate and generalization capability of a pre-trained model. 
We then construct the DAM via continuous relaxation and discretization with weight-sharing optimizations.
We fully implement DiffoRA and design comprehensive experiments to evaluate its performance. 
The experimental results demonstrate that DiffoRA delivers state-of-the-art results across multiple benchmarks.
\end{abstract}

\section{Introduction}
\label{sec:intro}

Large language models (LLMs) have gained significant attention across various domains of natural language processing, demonstrating remarkable capabilities in tasks such as text generation \cite{li2024dtllm, qu2023layoutllm}, machine translation \cite{xu2024contrastive}, sentiment analysis \cite{xing2024designing}, and question-answering \cite{zhuang2023toolqa}. 
%
Due to the large model size, fine-tuning an LLM to downstream tasks will invoke a large number of parameters, \eg, up to 175 billion parameters to fully fine-tune GPT-3.
As a result, Parameter-Efficient Fine-Tuning (PEFT) has become increasingly paramount.

The existing PEFT methods can be categorized into two classes.
Some approaches slightly modify the model structure by adding small trainable modules and keeping the rest of the model unchanged \cite{liu2022ptuning, lester2021power}. 
Another line of work aims to efficiently capture the incremental parameter updates without modifying the model structure \cite{guo2021parameter}. Among all the existing work, the Low-Rank Adaptation \cite{hu2022lora} is widely acknowledged due to its effectiveness and satisfying performance. Unlike the previous work, LoRA incurs some low-rank decomposition matrices to parameterize the incremental updates and realizes comparable results with 70\% less overhead than the full fine-tuning. This innovative approach paves the way for more effective utilization of LLMs in practice.

Although LoRA is effective in practice, it treats all the modules in the network equally by adding the same decomposition matrices to all the trainable modules in the network. 
To alleviate this issue, some work has been proposed as adaptive LoRA, which managed to adjust the interior ranks of decomposition matrices for each module based on parameter importance estimation \cite{zhang2024autolora, zhang2023adaptive, liu2024alora, mao2024doraadaptive, zhou2024loradrop}.
For example, \citet{zhang2023adaptive} utilizes the singular value of the low-rank matrices to adjust the rank, while  SalientLoRA \cite{NEURIPS2024_ed9f00cb} uses a salience measurement to adaptively optimize the ranks of LoRA.

%

\begin{figure}
    \centering
    \includegraphics[width=.9\textwidth]{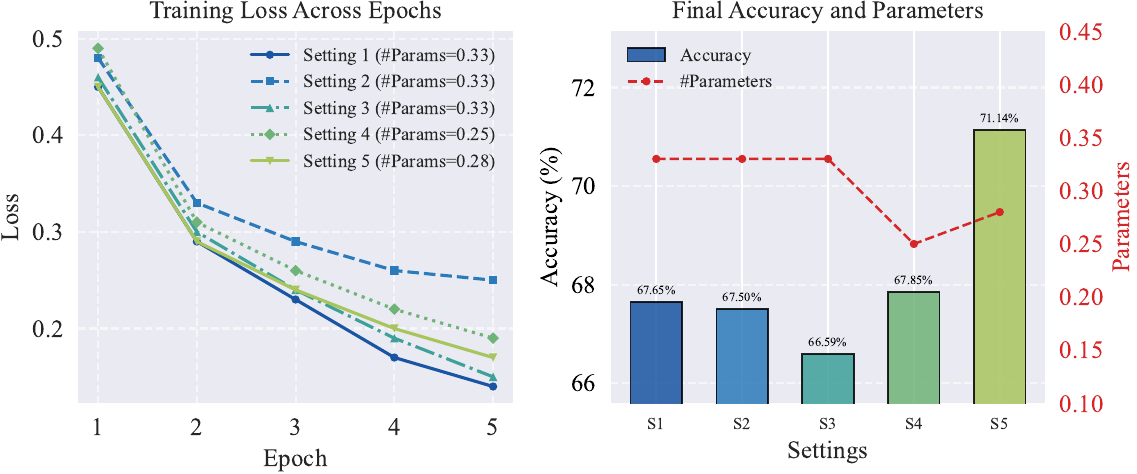}
    \caption{The loss (left) and final accuracy (right) obtained on the CoLA dataset and DeBERTaV3-base using LoRA under different settings. 
    }
    \label{fig: observe}
\end{figure}

Though effective, we point out that fine-tuning all modules, like the previous work, is not necessary and might inversely affect the model performance. 
As justification, we conduct a toy experiment by comparing the accuracy of DeBERTaV3-Base on CoLA \cite{wang2018glue} using LoRA when selectively fine-tuning: S1) all modules; S2) only the QKV projections; S3) only dense modules; S4) QKV projections in the first half of the model and dense modules in the remaining half; and S5) three modules per layer that are randomly selected. We repeat the selection for 10 times to find the optimal configuration.
It can be seen from the results in Figure \ref{fig: observe} that 
\begin{itemize}
    \item Different modules and combination strategies contribute differently to the fine-tuning results. An appropriate module combination can significantly enhance the model's performance. For instance, by randomly selecting the modules in each layer, S5 achieves over 4\% higher accuracy than other settings with lower parameter amounts.
    \item Not all of the modules are suitable and necessary to be fine-tuned. For instance, S1 obtains a lower accuracy than S4 and S5, which also leads to overfitting. Some modules in S1 contribute negatively to the performance, thus, it is unnecessary to adjust their interior ranks.  
\end{itemize}
The above observations demonstrate the importance of selecting the most essential modules during LoRA fine-tuning, which is often overlooked by the existing methods. This presents a compelling research question: 
\textit{Can we construct a theoretically grounded PEFT method to enable adaptive adoption of the low-rank decomposition matrices, so that we can only fine-tune the modules that are most necessary and essential? 
}



\noindent\textbf{Our contributions.} 
In this paper, we affirmatively answer the question with DiffoRA, a novel method to improve PEFT performance based on LoRA. 
Instead of adjusting the interior rank of every decomposition matrix, DiffoRA only adopts the low-rank matrices for those modules that are suitable and necessary to be fine-tuned.  
To this end, at the core of our approach lies a Differentiable Adaptation Matrix (DAM) that performs a module selection. 
DAM is designed to be differentiable with respect to the incremental updates of the low-rank matrices, allowing it to effectively capture fine-tuning priorities for various network backbones.
We provide a theoretical analysis of how DAM influences convergence rate and generalization during fine-tuning. 
Then, we approximate the NP-hard optimization by constructing DAM via continuous relaxation and discretization. 
To mitigate discretization-induced discrepancies and enhance robustness, we further introduce a weight-sharing strategy.
We implement DiffoRA and conduct extensive experiments on three standard benchmarks for both Natural Language Understanding and Question Answering tasks.
The results demonstrate that DiffoRA consistently outperforms existing methods. For example, it achieves a 1.51\% accuracy gain over the state-of-the-art on the CoLA dataset.
The main contributions of this work are summarized as follows:

\begin{itemize}
    \item We propose DiffoRA, a novel and theoretically-grounded PEFT method for LLMs. Our approach is built atop a newly designed differentiable matrix (\ie, DAM), which enables module-wise adoption of the low-rank decomposition matrices. 
    
    \item We theoretically explain how the DAM impacts the performance of a LoRA-based fine-tuning model in terms of the convergence rate and generalization capability.

    \item We fully implement DiffoRA and evaluate its performance on several widely adopted benchmarks that contain multiple tasks. The experimental results demonstrate that DiffoRA works consistently better than all the baselines.  
\end{itemize}

\section{Related work}
\label{sec: related work}

\subsection{Parameter-efficient fine-tuning.}
LLMs have captured considerable public interest due to their success in multiple regions. The PEFT is essential for LLMs due to the huge number of parameters.  
Some previous work has been proposed to fine-tune the LLMs using specifically designed modules that are added to the LLMs. The fine-tuning of LLMs is thus converted to the adjustments of these small modules. For instance, multiple methods  \cite{rebuffi2018efficient, liu2022ptuning, lester2021power, NEURIPS2024_4a9eaf6d} insert dataset-dependent small modules or vectors between the layers to decrease the parameter amounts. 
Another line of work models the incremental updates of the fine-tuning procedure to make it parameter-efficient. \citet{guo2021parameter} propose to use a task-specific difference vector to extend and fine-tune the base model. There are also some methods that fine-tune parts of 
the parameters \cite{gui2023hifi}, \eg, the bias of FFN \cite{zaken2022bitfit} or the last quarter \cite{lee2019would}. 

\subsection{Low-rank adaptation and optimizations}

To further reduce the computational and storage cost, \citet{hu2022lora} proposed LoRA, in which they designed a low-rank decomposition matrix to model the incremental updates during fine-tuning. 
A plethora of work has been proposed to optimize LoRA \cite{kopiczko2024vera, hayou2024, liu2024dora, dettmers2024qlora, renduchintala2024tied, hayou2024lora+, wang2025lorapro, tian2024hydralora}. One of the limitations of LoRA is that it treats all the modules equally, which omits the variations of the modules in LLMs. 
To address this issue, a few work has been proposed to realize an adaptive LoRA.
These methods first evaluate the importance of the modules and then adjust the interior rank of the decomposition matrices accordingly, e.g., AdaLoRA \cite{zhang2023adaptive} and SalientLoRA \cite{NEURIPS2024_ed9f00cb}.
There is also a few work that utilize the training procedures to determine the module importance \cite{ding2023sparse, zhang2024autolora, liu2024alora}, yet they still focus on modifying the interior ranks of the decomposition matrices.

\section{Theoretical analysis}
\subsection{Preliminaries}

\textbf{Notations.} We define $[n]=\{1,2,\ldots, n\}$. 
Vectors and matrices are represented by bold lowercase and bold uppercase fonts, respectively. For instance, $\bm{x}$ is a vector with entry $x_i$, and $\bm{M}$ is a matrix with entry $[\bm{M}]_{ij}$.
The minimum eigenvalue of $\bm{M}$ is denoted as $\lambda_{\min}(\bm{M})$.
$\|\cdot\|_2$ is used to represent the $l_2$ norm of a vector. 
$N(\bm{0},\bm{I})$ and $U\{S\}$ represent the standard Gaussian distribution and uniform distribution over a set $S$, respectively. 
%
We denote by $\bm{X}=\{(\bm{x}_i, y_i)|\bm{x}_i\in\mathbb{R}^{d\times1}, y_i\in\mathbb{R},i\in[n]\}$ the training set, where $\bm{x}_i$ and $y_i$ represent the $i$-th data and label.
$\mathbb{I}\{\cdot\}$ represents the indicator function that demonstrates the event occurrence, such that for event $\mathcal{A}$, $\mathbb{I}\{\mathcal{A}\}=1$ if and only if $\mathcal{A}$ happened, otherwise it equals 0. $P(\mathcal{A})$ represents the probability of $\mathcal{A}$ occurred event.

For input $\bm{x}\in \mathbb{R}^{d\times1}$, weight vector $\bm{w} \in \mathbb{R}^{d\times1}$ in the weight matrix $\bm{W} \in \mathbb{R}^{d\times m}$, and output weight $\bm{a} \in \mathbb{R}^{m\times 1}$, we denote $f(\bm{W},\bm{a},\bm{x})$ as a neural network with a single hidden layer such that
\begin{equation}\label{eq: neural}
f(\bm{W},\bm{a},\bm{x})=\frac{1}{\sqrt{m}}\sum\nolimits_{r=1}^m a_r\sigma(\bm{w}_r^T \bm{x})
\end{equation}
where $\sigma$ is the activation function. In this paper, we primarily consider the ReLU function, which is one of the most widely adopted activation functions in the literature, \ie, $\sigma(z)=z\mathbb{I}\{z>0\}$.
Given a training set $\bm{X}$, the optimization goal is to minimize the empirical risk loss function 
\begin{equation}\label{eq:loss}
\mathcal{L}(\bm{W},\bm{a})=\sum\nolimits_{i=1}^n\frac{1}{2}(f(\bm{W},\bm{a},\bm{x}_i)-y_i)^2
\end{equation}

\noindent\textbf{Recall LoRA.} \citet{hu2022lora} utilize two matrices $\bm{A} \in \mathbb{R}^{r \times k}, \bm{B} \in \mathbb{R}^{d \times r}$ to substitute the parameters' increments, where $d,k \ll r$, $\bm{A}$ is usually initialized by following Gaussian distribution and $\bm{B}$ is initialized with zeros. For $\bm{h}=\bm{W}^{0} \bm{x}$, the forward pass in LoRA is
\begin{equation}
    \bm{h} = \bm{W}^0 \bm{x} + \Delta \bm{x} = \bm{W}^0 \bm{x} + \bm{B}\cdot \bm{A} \bm{x} 
\end{equation}
where $\bm{W}^0$ denotes the weight matrix of the pre-trained network, consisting of $r$ row vectors $\bm{w}_{0,r}$. 
In the following, we denote $\Delta \bm{W}$ as $\bm{B}\cdot \bm{A}$.
LoRA adopts this modification equally to all the modules in the model.
Therefore, fine-tuning the pre-trained network using LoRA can be formulated as
\begin{equation}\label{eq: neurallora}
f(\bm{W},\bm{a},\bm{x};\bm{W}_0)=\frac{1}{\sqrt{m}}\sum\nolimits_{r=1}^m a_r\sigma((\bm{w}_{0, r}+\bm{w}_r)^T \bm{x})
\end{equation}
where $\bm{w}_r$ is the $r$-th row of the trainable parameters in the low-rank matrices.
%


\subsection{Main theorems}
\noindent\textbf{Technical intuitions.} In this work, we aim to construct a matrix called DAM to determine the necessity of each module for fine-tuning in LLM.
We denote DAM as $\bm{\Gamma}$. 
Intuitively, $\mathbf{\bm{\Gamma}}$ is applied to the trainable parameters of the LoRA module-wise, such that the module corresponding to $\gamma_{i,j}=1$ is determined to be fine-tuned, and vice versa.
%
%
Notably, the module-wise fine-tuning can be viewed as a special case of an element-wise formulation, where entire submatrices are either enabled or disabled uniformly.
For analytical clarity, we focus on the more general element-wise setting, which allows finer-grained control and provides deeper insights into how the structure of $\bm{\Gamma}$ influences models' performance.
In what follows, we assume access to a suitable algorithm for constructing $\bm{\Gamma}$ and aim to elucidate how such selective adaptation contributes to improved model performance.
To facilitate this analysis, we extend the definitions introduced in \cref{eq: neural} as follows:
\begin{equation}\label{eq: lora}
    \begin{aligned}
f(\bm{W},\bm{a},\bm{x};\bm{\Gamma},\bm{W}_0)=\frac{1}{\sqrt{m}}\sum_{r=1}^m a_r\sigma((\bm{w}_{0,r}+\bm{\gamma}_r\circ\bm{w}_r)^T\bm{x}) 
    \end{aligned}
\end{equation}
where $\bm{\gamma}_r$ is the $r$-th row of $\bm{\Gamma}$ and $\circ$ represents the Hadamard product.
Furthermore, we follow the definitions in \cite{dugradient} and define $\mathbf{H}^\infty_{\bm{\Gamma}, \bm{w}_0}$ and $\mathbf{H}^\infty_{\bm{w}_0}$ based on $\bm{\Gamma}$ such that
\begin{definition}[Gram Matrix with $\bm{\Gamma}$]\label{def: gram}
For a neural network with a single hidden layer and ReLU activation, the Gram matrix $\bm{H}^\infty_{\bm{\Gamma}, \bm{w}_0} \in \mathbb{R}^{n \times n}$ induced on a training set $\bm{X}:= \{(\bm{x}_i, y_i)\}_{i=1}^{n}$ is defined entry-wise as
\begin{equation}\label{eq: gram}
\begin{aligned}\relax
[\bm{H}_{\bm{\Gamma},\bm{w}_0}^\infty]_{ij}
=&\mathbb{E}_{\bm{w}\sim N(\bm{0},\bm{I}), r\in[m]}[\bm{x}_i^T\bm{x}_j\cdot \mathbb{I}\{(\bm{w}_{0,r}+\bm{\gamma}_r\circ\bm{w})^T\bm{x}_i\geq 0,(\bm{w}_{0,r}+\bm{\gamma}_r\circ\bm{w})^T\bm{x}_j\geq 0\}]\\
=&\bm{x}_i^T\bm{x}_j[\bm{I}^{\bm{\Gamma}\bm{w}}]_{ij}
\end{aligned}
\end{equation}

We also construct $\mathbf{H}_{\bm{w}_0}^\infty$ with entry $[\mathbf{H}_{\bm{w}_0}^\infty]_{ij}$ such that 
\begin{equation}
\begin{aligned}\relax
    [\bm{H}^\infty_{\bm{w}_0}]_{ij} =& \mathbb{E}_{\bm{w}\sim N(\bm{0},\bm{I}), r\in[m]}[\bm{x}_i^T\bm{x}_j\cdot \mathbb{I}\{(\bm{w}_{0,r}+\bm{w})^T\bm{x}_i\geq 0,(\bm{w}_{0,r}+\bm{w})^T\bm{x}_j\geq 0\}]\\
    =&\bm{x}_i^T\bm{x}_j[\bm{I}^{\bm{w}}]_{ij}
\end{aligned}
\end{equation}
where $\bm{I}^{\bm{\bm{\Gamma} w}}$ and $\bm{I}^{\bm{w}}$ are the expectations of the indicator matrices.
We denote $\lambda_0 := \lambda_{\min}(\mathbf{H}_{\bm{w}_0}^\infty)$, and $\lambda_0^{\bm{\Gamma}}:=\lambda_{\min}(\mathbf{H}_{\bm{\Gamma},\bm{w}_0}^\infty)$.
\end{definition}

\noindent\textbf{Theoretical results.} 
The theoretical analysis is based on the intuitive observation that fine-tuning a pre-trained model can be viewed as training a well-initialized model.
We analyze LoRA fine-tuning from the perspective of over-parameterized neural networks, focusing on two fundamental aspects: the convergence rate and the generalization capability.
Over-parameterized networks have been widely adopted and demonstrated strong empirical performance in hierarchical feature extraction, primarily due to their large capacity and training tractability \cite{li2018visualizing, jiang2024meco}.
In this section, we leverage this architecture to establish our theoretical understanding of the matrix $\mathbf{\bm{\Gamma}}$.
Building on prior work \cite{dugradient}, we present the following theorem that characterizes the convergence behavior of a single-hidden-layer neural network in the over-parameterized setting.
We first analyze the relationship of the minimum eigenvalue between the adaptive matrix (\ie, $\lambda^{\bm{\Gamma}}_0$) and the original matrix (\ie, $\lambda_0$). 
We summarize our results in the following theorem: 

\begin{theorem}\label{theorem:convergence}
    Suppose $f$ is an NN with a single hidden layer and ReLU activation function. 
    Assume $\bm{X}\in \mathbb{R}^{d\times n}$, $\bm{w}(0)\sim N(\bm{0},\bm{I})$, hidden nodes $m=\Omega\left(\frac{n^6d^2}{(\lambda_0^{\bm{\Gamma}})^4\delta^3}\right)$, and $\bm{I}^{\bm{\bm{\Gamma} w}}-\bm{I}^{\bm{w}}\succeq 0$, then the following formula holds with probability at least $1-\delta$ over the initialization
    \begin{equation}
    \begin{aligned}
        \|f(\bm{W}(t),\bm{a},\bm{X};\bm{\Gamma},\bm{W}_0)-\bm{y}\|_2^2 &\leq \exp(-\lambda_0^{\bm{\Gamma}} t)\|f(\bm{W}(0),\bm{a},\bm{X};\bm{\Gamma},\bm{W}_0)-\bm{y}\|_2^2\\
        &\leq\exp(-\lambda_0 t)\|f(\bm{W}(0),\bm{a},\bm{X};\bm{\Gamma},\bm{W}_0)-\bm{y}\|_2^2
    \end{aligned}
    \end{equation}
    where $\lambda_0^{\bm{\Gamma}}\geq \lambda_0$. 
\end{theorem}

\emph{Proof sketch.} The key to the proof lies in identifying relationship between $\lambda_0$ and $\lambda_0^{\bm{\Gamma}}$ by Weyl inequalities \cite{horn2012matrix}. We provide the full proof in \cref{app:sub:th2}. 

%
Theorem \ref{theorem:convergence} demonstrates that the minimum eigenvalue of the Gram matrix of the network with $\mathbf{\bm{\Gamma}}$ is larger than the $\lambda_0$ (without the selective matrix), thus leading to a higher convergence rate. 
%
%
Other than the convergence rate, we also analyze the relationship between $\lambda_0^{\bm{\Gamma}}$ and the generalization capability of the model. We present the results in the following theorem:

\begin{theorem}\label{theorem: genera}
    For an over-parameterized neural network with the loss on the testing set as $\mathcal{L}(\bm{W}, \bm{a};\bm{\Gamma},\bm{W}_0)$. Let $\bm{y}=(y_1, ..., y_N)^T$, and $\eta = \kappa C\sqrt{\bm{y}^T(\bm{H}_{\bm{\Gamma},\bm{w}_0}^\infty)^{-1}\bm{y}}/(m\sqrt N)$ for some small enough absolute constant $\kappa$, where $\eta$ denotes the step of SGD. Under the assumption of Theorem \ref{theorem:convergence}, for any $\delta\in(0,e^{-1}]$, there exists $m^\ast(\delta,N,\lambda_0^{\bm{\Gamma}})$, such that if $m\geq m^*$, then with probability at least $1-\delta$, we have
    \begin{equation}\label{eq: generalization}
    \begin{aligned}
        \mathbb{E}[\mathcal{L}(\bm{W},\bm{a};\bm{\Gamma},\bm{W}_0)]&\leq  \mathcal{O}(C'\sqrt{\frac{\bm{y}^T \bm{y}}{\lambda_0^{\bm{\Gamma}} N}})+ \mathcal{O}(\sqrt{\frac{\log(1/\delta)}{N}})
        \leq  \mathcal{O}(C'\sqrt{\frac{\bm{y}^T \bm{y}}{\lambda_0 N}})+ \mathcal{O}(\sqrt{\frac{\log(1/\delta)}{N}})
    \end{aligned}
    \end{equation}
    where $\mathcal{L}(\bm{W},\bm{a};\bm{\Gamma},\bm{W}_0)$ is the loss corresponding to \cref{eq: lora}, $C, C'$, and $\delta$ are constants. 
\end{theorem}

\emph{Proof sketch.} The proof of the above theorem derives from Corollary 3.10 of \cite{cao2019generalization} and Section D.2 of \cite{zhu2022generalization}. We present the detailed proof in \cref{app:sub:th3}. 

Similarly to Theorem \ref{theorem:convergence}, the above theorem indicates that the adoption of $\bm{\Gamma}$ can enhance the generalization capability of the model, as shown in the last inequality of \cref{eq: generalization}.

\begin{remark}
In this section, we theoretically demonstrate that incorporating the DAM matrix $\bm{\Gamma}$ into a LoRA-based network can improve both the convergence rate and generalization performance, ultimately leading to enhanced model effectiveness. Theorem \ref{theorem:convergence} and \ref{theorem: genera} provide formal justifications for the benefits of $\bm{\Gamma}$. The remaining challenge lies in how to efficiently construct such a matrix. We address this in the following section, where we introduce practical solutions and detailed algorithms. 
\end{remark} 


\section{Design of DiffoRA}

\begin{figure}
    \centering
    \includegraphics[width=\textwidth]{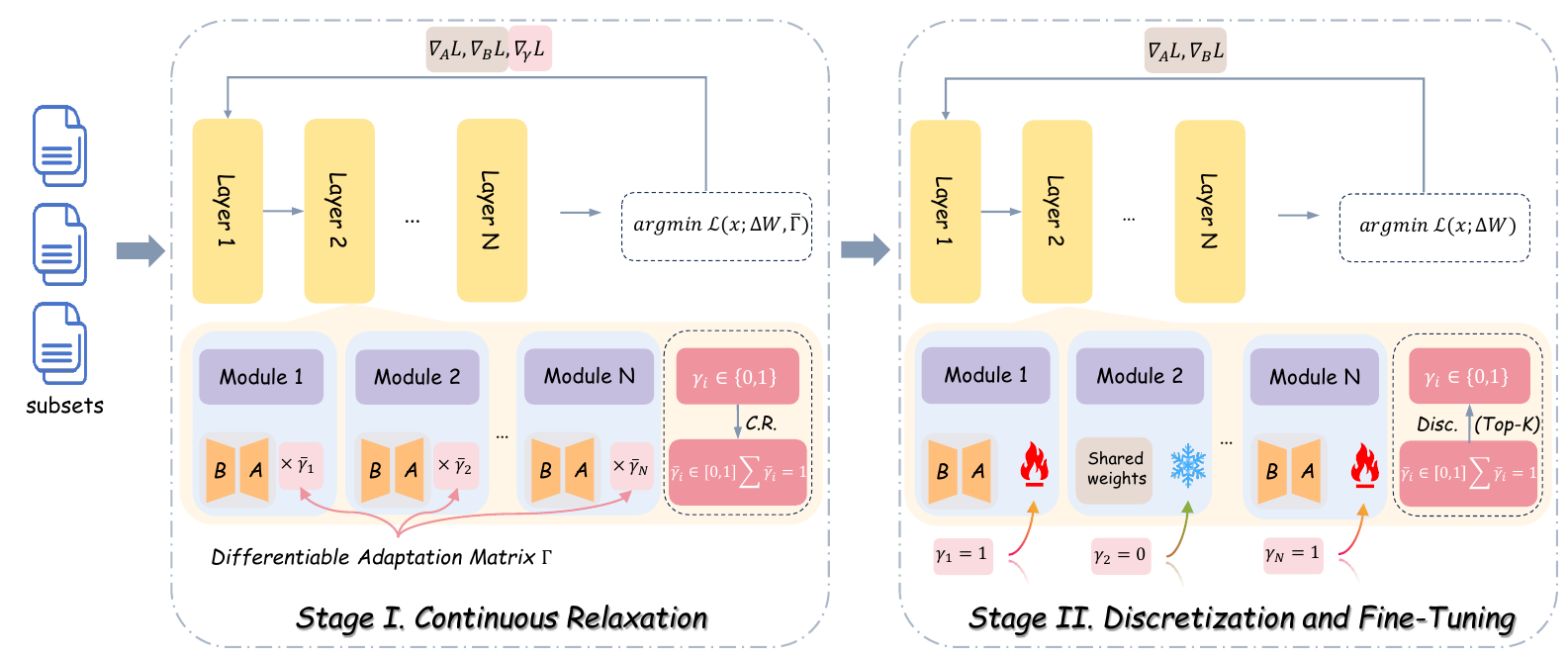}
    \caption{The framework of DiffoRA contains two stages. In stage one (left part), the initialized adaptive matrix is continuously relaxed, \ie, C.R. in the figure. The DAM is then differentiable and can be updated. In stage two (right part), the obtained DAM is discretized into binary. All the modules corresponding to entry one will be fine-tuned using decomposition matrices.}
    \label{fig: difforaframe}
\end{figure}

We now present the concrete method to construct the module-wise selective matrix $\mathbf{\bm{\Gamma}}$. The proposed approach is called DiffoRA, which is built atop the DAM (\ie, \underline{D}ifferentiable \underline{A}daptation \underline{M}atrix).
%
%
To capture the information of the incremental updates modeled by the low-rank decomposition matrices, DiffoRA views the elements $\gamma$ in $\bm{\Gamma}$ as trainable parameters.
More concretely, as shown in \cref{fig: difforaframe}, DiffoRA approximates the NP-hard problem by invoking the following two stages: (i) Relaxation and optimization, which aims to map $\bm{\Gamma}$ to a continuous space (\ie, $\bar{\bm{\Gamma}}$) so that it is differentiable and can be further optimized; and (ii) Discretization and fine-tuning, which binarizes $\bar{\bm{\Gamma}}$ to determine the most essential module for fine-tuning. We will first elaborate on each stage, respectively, and then introduce the optimization method called weight-sharing.



\subsection{Continuous relaxation}

The first stage of DiffoRA is illustrated in the left part of \cref{fig: difforaframe}.
In contrast with the existing work, we point out that it is unnecessary to allocate ranks for every module. 
Instead, we can construct a selective matrix called DAM that determines which module needs to be fine-tuned and which does not. 
The final output $\bm{\Gamma}$ is a binary matrix, in which the ``ones'' (\textit{resp.} ``zeros'') indicate that the corresponding entries will (\textit{resp.} will not) be fine-tuned. 
However, directly generating a binary matrix is non-trivial, \ie, it is an NP-hard problem.
To address this challenge, DiffoRA relaxes the binary constraint on the DAM matrix from $\bm{\Gamma} \in \{0, 1\}^{L \times N}$ to a continuous form $\bar{\bm{\Gamma}} \in [0, 1]^{L \times N}$, where $L$ and $N$ denote the number of model layers and the number of trainable modules per layer, respectively. All row vectors in $\bar{\bm{\Gamma}}$ are treated as differentiable hyperparameters, allowing them to be jointly optimized with the network via gradient-based methods.


More specifically, 
for the collection of the modules in the $i$-th layer, we utilize the row vector $\bar{\bm{\gamma}}_i\in [0,1]^{1\times N}$ of $\bm{\bar{\bm{\Gamma}}}$ as learnable hyperparameters in continuous space.
The forward pass of the relaxed LoRA-based fine-tuning formula can then be defined as:
\begin{equation}\label{con-relax}
    h_{i,j}=\bm{W}_{i,j} \bm{x}+\bar{\gamma}_{i,j} \Delta \bm{W}_{i,j} \bm{x}=\bm{W}_{i,j} \bm{x}+\bar{\gamma}_{i,j} \bm{B}_{i,j} \bm{A}_{i,j} \bm{x},
\end{equation}
where $h_{i,j}$ denotes the $j$-th item of the hidden nodes in the $i$-th layer, and each $\bar{\gamma}_{i,j}$ in $\bar{\bm{\gamma}}_i$ satisfies $\sum_{j}\bar{\gamma}_{i,j}=1$, and $\bar{\gamma}_{i,j}\geq 0$.
%
$\bm{W}_{i,j}\in\mathbb{R}^{d\times k}$ is the pre-trained weight matrix of module $m_{i,j}$.
$\bm{B}_{i,j}$ and $\bm{A}_{i,j}$ are the $j$-th low-rank decomposition matrices in the $i$-th layer.

Intuitively, $\bm{\bar{\bm{\Gamma}}}$ can represent the necessities of fine-tuning this module using $\mathbf{A}$ and $\mathbf{B}$. 
When this weight tends to $1$, it indicates that fine-tuning this module is necessary and might lead to better performance, and vice versa. 
By utilizing the continuous relaxation on the differential matrix, we are able to generate each $\bar{\bm{\gamma}}_i$ via a few rounds of training (\eg, five rounds, determined by the datasets).  
%
We denote $\mathcal{L}_{train}$ and $\mathcal{L}_{valid}$ as the training and validation loss.
%
Then, our goal is to find the best hyperparameters $\bar{\bm{\Gamma}}$ to minimize the validation loss, which is equivalent to bi-level optimization problems as follows:
\begin{equation}\label{eq:opt}
    \begin{aligned}
        \min_{\bar{\bm{\Gamma}}} \quad&\mathcal{L}_{valid}( \Delta \bm{W}^\ast, \bar{\bm{\Gamma}})\\
        s.t.\quad &\Delta \bm{W} ^{\ast}=\mathop{\arg\min}_{\Delta \bm{W}}\mathcal{L}_{train}(\Delta \bm{W}, \bar{\bm{\Gamma}})
    \end{aligned}
\end{equation}
To solve \cref{eq:opt}, we use the gradient descent algorithm to update the parameters $\bm{A}$, $\bm{B}$ and $\bar{\bm{\Gamma}}$. 
Specifically, we randomly extract part of the data in the training set and divide it into two parts, \ie, the training data and validation data, which are used to update $\Delta \bm{W}$ and $\bar{\bm{\Gamma}}$ alternately. We present the detailed training algorithms in \cref{alg: difflora} in Appendix \ref{app:alg}, lines 1 to 8.

\subsection{Discretization and fine-tuning}

Having the relaxed matrix $\bar{\bm{\Gamma}}$ in the previous stage, we perform the discretization to obtain the binary DAM $\bm{\Gamma}$. 
An intuitive representation of this procedure is shown in the right part of \cref{fig: difforaframe}.
More concretely, for each item in $\bar{\bm{\Gamma}}$, we update the top $K := \lfloor\rho \cdot N\rfloor$ largest entries with 1 and set the remaining values as 0, such that
\begin{equation}
    \gamma_{i,j}=
   \begin{cases}
   1 & \mbox{if $\bar{\gamma}_{i,j} \ge \delta_i$}\\
   0 & \mbox{otherwise}
   \end{cases}
\end{equation}
where $\delta_i$ is the value of the $K$-th largest entry in $\bar{\bm{\gamma}}_i$ and $\rho$ is the selection ratio.
The discretization is described in \cref{alg: difflora}, lines 10 and 11.
The final step is to fine-tune the model equipped with the binarized DAM and low-rank decomposition matrices. 
During the fine-tuning, only the modules with weight one (\eg, the selected module) will be fine-tuned, while the others are kept unchanged.


\subsection{Optimization with weight sharing}


\begin{figure}
    \centering
    \includegraphics[width=\textwidth]{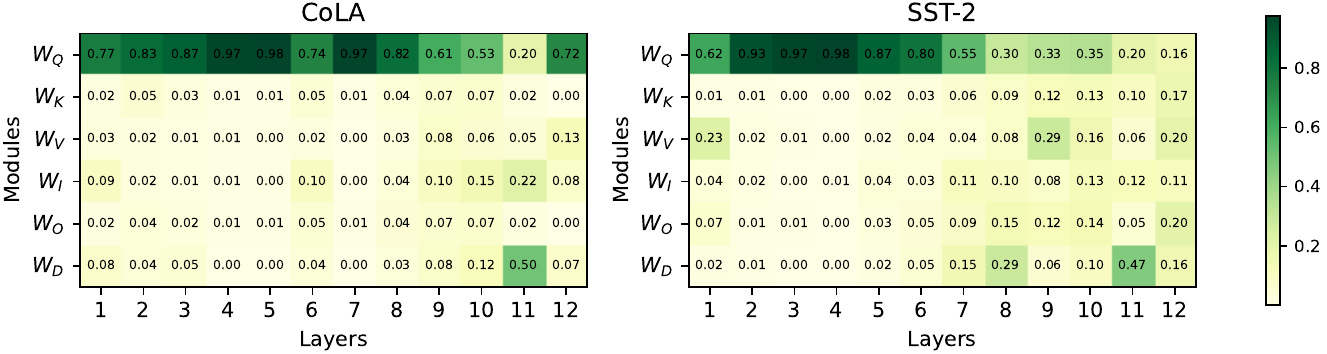}
    \caption{The module weights of the DeBERTaV3-base model in the CoLA and SST-2 datasets. 
    }
    \label{fig: module weight}
\end{figure}


In the previous section, we constructed the DAM $\bm{\Gamma}$ via continuous relaxation and discretization. While this method significantly improves fine-tuning performance when the relaxed weights $\bar{\bm{\Gamma}}$ are highly discriminative, its effectiveness diminishes under uniformly distributed weights.
As shown in \cref{fig: module weight}, we visualize $\bar{\bm{\Gamma}}$ on CoLA and SST-2 \cite{wang2018glue} using DeBERTaV3-base \cite{he2023debertav} as the backbone, where each column represents a layer with six candidate modules. 
On CoLA, weight distributions are relatively distinct—e.g., $W_Q$ dominates in most layers, facilitating effective module selection. In contrast, SST-2 exhibits nearly uniform weights, with all modules receiving values around 0.15 across layers 8–12. In such cases, top-$K$ selection may discard modules that are equally important, leading to potential performance degradation if zero-valued modules in $\bm{\Gamma}$ are entirely frozen.

To address this issue, we introduce a weight-sharing optimization strategy. 
For insufficiently discriminative DAM, we do not exclude modules with zero entries from the fine-tuning process, but allow them to share weights with corresponding modules in different layers. 
Specifically, all unselected modules $\bm{W}_i \in \{\bm{W}_Q, \bm{W}_K, \bm{W}_V, \bm{W}_I, \bm{W}_O, \bm{W}_D\}$ for all layers share the same set of weights and are jointly updated during fine-tuning. This strategy is particularly beneficial for handling DAMs with ambiguous distinctions, enabling modules with low discriminability to still participate in optimization. As a result, DiffoRA's effectiveness is enhanced without introducing additional parameter overhead. The corresponding implementation details are provided in \cref{alg: difflora}, lines 12–15.

\section{Experiments}\label{sec: experiments}

\subsection{Configurations}\label{subsec: configurations}



\noindent\textbf{Datasets and pre-trained models.} 
We utilize two types of benchmarks: i) General Natural Language Understanding (GLUE) \cite{wang-etal-2018-glue}, including MNLI, SST-2, CoLA, QQP, QNLI, RTE, MRPC, and STS-B; and ii) Question Answering, including SQuADv1.1 \cite{rajpurkar-etal-2016-squad} and SQuADv2.0 \cite{rajpurkar-etal-2018-know}. In the main text, we adopt DeBERTaV3-base \cite{he2023debertav}, LLaMA-3.2-1B \cite{grattafiori2024llama}, and LLaMA-7B \cite{touvron2023llama} as backbone models. Additionally, results obtained on the GLUE benchmark using RoBERTa-base \cite{liu2019roberta} are provided in \cref{app:roberta} for comparison.

\noindent\textbf{Counterpart comparisons.} We choose representative LoRA and adaptive LoRA methods from the literature as baselines, ensuring comparable parameter budgets and availability of open-source implementations. 
The details can be found in \cref{app:baseline}. Our code is available at the \href{https://anonymous.4open.science/r/DiffoRA-2128}{https://anonymous.4open.science/r/DiffoRA-2128}.


\begin{table}
  \centering
    \caption{The results of the fine-tuned DeBERTaV3-base model on the GLUE dataset are presented, with the best results highlighted in bold and the second-best results underlined. 
    }
  \label{tab: glue}
  \resizebox{\textwidth}{!}{
  \begin{tabular}{@{}l|c|ccccccccc@{}}
    \toprule
    \multirow{2}{*}{Method} & \multirow{2}{*}{\#Params} & MNLI & SST-2 & CoLA & QQP & QNLI & RTE & MRPC & STS-B & All \\
     & & m/mm & Acc & Mcc & Acc/F1 & Acc & Acc & Acc & Corr & Avg.\\
    \midrule
    Full FT & 184M & 89.90/90.12 & 95.63 & 69.19 & 92.40/89.80 & 94.03 & 83.75 & 89.46 & 91.60 &88.09\\
    \midrule
    BitFit & 0.1M & 89.37/89.91 & 94.84 & 66.96 & 88.41/84.95 & 92.24 & 78.70 & 87.75 & 91.35 & 86.02 \\
    \midrule
    HAdapter &0.61M& 90.12/90.23& 95.30& 67.87 &91.65/88.95& 93.76& 85.56 &89.22& 91.30& 87.93\\
    PAdapter& 0.60M& 90.15/90.28& 95.53 &69.48 &91.62/88.86& 93.98 &84.12 &89.22& 91.52& 88.04\\
    HAdapter &0.31M& 90.10/90.02& 95.41& 67.65& 91.54/88.81& 93.52& 83.39& 89.25& 91.31& 87.60\\
    PAdapter &0.30M& 89.89/90.06& 94.72 &69.06 &91.40/88.62& 93.87& 84.48& 89.71 &91.38& 87.90\\
    LoRA$_{r=2}$ &0.33M& 90.30/90.38& 94.95 &68.71& 91.61/88.91 &94.03 &85.56& 89.71& \underline{91.68}& 88.15\\
    AdaLoRA& 0.32M& \textbf{90.66}/\textbf{90.70} & \underline{95.80}& \underline{70.04}& \underline{91.78}/\underline{89.16}& \underline{94.49} &\underline{87.36} &\underline{90.44} &91.63& \underline{88.86}\\
    \midrule
 \multirow{2}{*}{DiffoRA}& \multirow{2}{*}{0.35M} &\cellcolor{gray!20}  \cellcolor{gray!20}\underline{90.55}/\underline{90.55} &\cellcolor{gray!20}\textbf{96.09} &\cellcolor{gray!20}\textbf{71.55} &\cellcolor{gray!20} \textbf{92.07}/\textbf{89.46}&\cellcolor{gray!20} \textbf{94.59}&\cellcolor{gray!20} \textbf{88.57}& \cellcolor{gray!20}\textbf{91.42}& \cellcolor{gray!20}\textbf{91.86}& \cellcolor{gray!20}{\textbf{89.66}}\\
    & &\cellcolor{gray!20} $\pm0.01$/$\pm0.01$ &\cellcolor{gray!20} $\pm 0.00$ &\cellcolor{gray!20} $\pm1.33$ &\cellcolor{gray!20} $\pm0.02$/$\pm0.03$ &\cellcolor{gray!20} $\pm0.08$ &\cellcolor{gray!20} $\pm0.52$ &\cellcolor{gray!20} $\pm0.88$ &\cellcolor{gray!20} $\pm0.04$&\cellcolor{gray!20}$\uparrow0.80$ \\
    \bottomrule
  \end{tabular}
  }

\end{table}

\begin{table}
  \centering
    \caption{The results of the fine-tuned DeBERTaV3-base model on the SQuAD dataset. We report EM/F1. The best results are highlighted in bold and the second-best results are underlined.}
      \label{tab: qa}
 \resizebox{\textwidth}{!}{
  \begin{tabular}{@{}l|cccc|cccc@{}}
    \toprule
    Method & \multicolumn{4}{c|}{SQuADv1.1} & \multicolumn{4}{c}{SQuADv2.0}\\
    
    \midrule
     Full FT &  \multicolumn{4}{c|}{86.0/92.7} &  \multicolumn{4}{c}{85.4/88.4}\\
     \midrule
    \#Params &0.08\% &0.16\%& 0.32\% & 0.65\% &0.08\% &0.16\%& 0.32\% & 0.65\%\\
    \midrule
    HAdapter&84.4/91.5&85.3/92.1&86.1/92.7& 86.7/92.9& 83.4/86.6& 84.3/87.3 & \underline{84.9/87.9} & \underline{85.4/88.3}\\
    PAdapter&84.4/91.7 &85.9/92.5&86.2/92.8& 86.6/93.0 &\textbf{84.2}/\textbf{87.2}& 84.5/\underline{87.6} & \underline{84.9}/87.8& 84.5/87.5\\
    LoRA&86.4/92.8 &86.6/92.9&86.7/93.1 &86.7/93.1&83.0/86.3 &83.6/86.7&84.5/87.4 &85.0/88.0\\
    AdaLoRA&\underline{87.2/93.4}& \underline{87.5/93.6}& \underline{87.5/93.7}& \underline{87.6/93.7} & 83.0/86.3& \underline{84.6}/87.5&84.1/87.3 &84.2/87.3\\
    \midrule
    \rowcolor{gray!20} DiffoRA &\textbf{87.6}/\textbf{93.5}&\textbf{88.1}/\textbf{93.8}& \textbf{88.1}/\textbf{93.8} & \textbf{88.1}/\textbf{93.9} & \textbf{84.2}/\textbf{87.2}&\textbf{84.8}/\textbf{87.8}&\textbf{85.1}/\textbf{88.0}&\textbf{85.5}\textbf{/88.4}\\
    \bottomrule
  \end{tabular}
  }

\end{table}

\subsection{Natural language understanding}
We evaluate the performance of the fine-tuned model on the GLUE benchmark \cite{wang2018glue}.
We use DeBERTaV3-base as the pre-trained model, which contains 183 million parameters. 
During this task, we set the rank of the low-rank decomposition matrices to 4 and fine-tuned 50\% of the modules in each layer.
We adopt all eight datasets in GLUE, and the concrete fine-tuning model architectures are determined by the specific tasks. 
We summarize and present the results in \cref{tab: glue}.
Overall, our DiffoRA model achieves the highest accuracy among all baselines at the same parameter count.
More specifically, on the CoLA dataset, our method outperforms AdaLoRA by 1.51\% in fine-tuning accuracy, and 0.8\% on average across all tasks.
In addition, we observe that DiffoRA consistently outperforms LoRA and the adaptive LoRA method, suggesting that some modules do not need to be fine-tuned, which is consistent with our observations.

\subsection{Question answering}
For the question-answering task, we fine-tune DeBERTaV3-base on SQuADv1.1 and SQuADv2.0 with varying parameter budgets. To ensure comparability with baselines, we set the LoRA rank to ${1, 2, 5, 10}$ for SQuADv1.1 and ${2, 4, 8, 15}$ for SQuADv2.0. Evaluation is conducted using Exact Match (EM) and F1, with results summarized in \cref{tab: qa}. DiffoRA consistently outperforms all baselines across both datasets. On SQuADv1.1, it achieves 0.4–0.5\% higher EM and 0.1–0.2\% higher F1 than AdaLoRA. Remarkably, it also surpasses full fine-tuning while updating only 0.08\% of the parameters. Similar gains are observed on SQuADv2.0, where DiffoRA outperforms all baselines and even full fine-tuning, further validating its effectiveness.

\subsection{Incorporating DiffoRA with existing methods}

To show the effectivenss of DiffoRA, we further incorporate it with LoRA+ \cite{hayou2024lora+}, which is an optimized LoRA method. We evaluate the performance on LLaMA-3.2-1B and LLaMA-7B across three NLU benchmarks: MNLI, SST-2, and QQP.. As shown in \cref{tab: llama-nlu}, both DiffoRA\textsubscript{LoRA} and DiffoRA\textsubscript{LoRA+} consistently outperform their counterparts. For example, on MNLI, DiffoRA improves the accuracy of LLaMA-3.2-1B by 2.25\% and 2.61\% over LoRA and LoRA+, respectively, while using fewer tunable parameters. These results confirm the superiority of DiffoRA over both standard and optimized LoRA variants across model scales.

In summary, the experimental results presented in this section demonstrate that DiffoRA consistently outperforms all baseline methods across various benchmarks and datasets. This highlights the importance of selecting the right modules for fine-tuning and aligns well with our theoretical insights.

\begin{table}
  \centering
    \caption{The results of the fine-tuned LLaMA model on the MNLI, SST-2 and QQP datasets are presented, with the best results highlighted in bold and the second-best results underlined.
    }
  \label{tab: llama-nlu}
    \resizebox{\textwidth}{!}{
  \begin{tabular}{c|c|c|cccc}
    \toprule
   Model& Method  &\#Params (\%)&MNLI (Acc)&SST-2 (Acc)& QQP (Acc/F1) & All (Avg.)\\
    \midrule
    \multirow{4}{*}{LLaMA-3.2-1B} & LoRA & 0.45 & 83.94& 95.30&  \underline{90.39}/\underline{86.89} & 89.13\\
    & LoRA+ & 0.45  &84.80 & 92.43& 88.21/84.19& 87.40 \\
    
    & \cellcolor{gray!20}DiffoRA$_{\text{LoRA}}$ &\cellcolor{gray!20}0.40&\cellcolor{gray!20}\underline{86.19} & \cellcolor{gray!20}\textbf{95.64}&\cellcolor{gray!20} 90.15/86.86 &\cellcolor{gray!20}\underline{89.71}\\
    
    & \cellcolor{gray!20}DiffoRA$_{\text{LoRA+}}$&\cellcolor{gray!20}0.40&\cellcolor{gray!20}\textbf{87.41}&\cellcolor{gray!20}\underline{95.07}&\cellcolor{gray!20}  \textbf{90.41}/\textbf{87.19}& \cellcolor{gray!20}\textbf{90.02}\\
    
        \midrule
    \multirow{4}{*}{LLaMA-7B} & LoRA & 0.30 & \underline{90.74}& \textbf{97.02}&  90.75/87.22 &91.43 \\
    
    & LoRA+ & 0.30 & 89.71 & 95.99& \underline{90.88}/87.67 &91.06\\
    
    &\cellcolor{gray!20} DiffoRA$_{\text{LoRA}}$ &\cellcolor{gray!20}0.22 &\cellcolor{gray!20}\textbf{90.79}& \cellcolor{gray!20}\textbf{97.02} &\cellcolor{gray!20}90.60/\underline{87.76}&\cellcolor{gray!20}\underline{91.54}\\
    
    &\cellcolor{gray!20}DiffoRA$_{\text{LoRA+}}$ &\cellcolor{gray!20}0.22&\cellcolor{gray!20}90.65 &\cellcolor{gray!20}\underline{96.56}&\cellcolor{gray!20} \textbf{91.30}/\textbf{88.51}&\cellcolor{gray!20}\textbf{91.76}\\
    \bottomrule
  \end{tabular}
  }

\end{table}



\section{Further analysis}

\noindent\textbf{DiffoRA vs. Random selection strategy.}
We analyze the effectiveness of DiffoRA by comparing its performance with that of the random selection method. 
Specifically, we randomly select three modules in each layer as the fine-tuning modules, and the remaining modules are processed with the weight-sharing strategy.
The results are shown in \cref{tab: random vs diffora}.
We choose ranks from $\{1, 2, 4\}$ for fine-tuning in the STS-B and SQuADv1.1 datasets, respectively.
It can be demonstrated that compared to random sampling, DiffoRA achieves significant performance improvements and effectively identifies important modules, e.g., DiffoRA is 0.2\% to 0.6\% higher than the random select strategy.  




\noindent\textbf{Sample rate. } 
We study the effects of different sample rates on DiffoRA. 
To explore the relationship between the performance and sample rate, we conduct experiments on three datasets: STS-B, RTE, and SQuADv1.1. 
We select sample rates from \{0.2, 0.4, 0.5, 0.7, 0.9\} corresponding to the most important Top-$\{1, 2, 3, 4, 5\}$ modules, respectively. 
The results are summarized in \cref{tab: sample rate}.
The results show that when the sampling rate is around 0.5, the performance of the fine-tuned model achieves state-of-the-art.
%


\begin{table}[htbp]
  \centering
  \begin{minipage}{0.45\textwidth}
    \centering
    \caption{Random Selection vs. DiffoRA.}
    \label{tab: random vs diffora}
     \resizebox{\textwidth}{!}{
    \begin{tabular}{c|c|c|c|c}
    \toprule
    Method  &Rank&\#Params&STS-B& SQuADv1.1\\
    \midrule
    Random & 1 & 0.07& 91.23&  87.23  \\
    \rowcolor{gray!20}DiffoRA & 1 & 0.07& 91.38 & 87.37 \\
    \midrule
    Random & 2 &0.14  & 91.49& 87.55  \\
    \rowcolor{gray!20}DiffoRA & 2 &0.13 & 91.81 & 87.74 \\
    \midrule
    Random &4 &0.27 & 90.99 & 87.82 \\
    \rowcolor{gray!20}DiffoRA & 4&0.34 & 91.86&  88.04 \\
    \bottomrule
  \end{tabular}
  }
  \end{minipage}
  \hspace{0.05\textwidth}
  \begin{minipage}{0.45\textwidth}
    \centering
    \caption{DiffoRA across three datasets at different sample rates.}
    \label{tab: sample rate}
    \resizebox{\textwidth}{!}{
\begin{tabular}{c|c|c|c|c}
    \toprule
    Sample Rate&K&STS-B& RTE & SQuADv1.1\\
    \midrule
    0.2  &1& 91.34 &87.73 & 87.09 \\
    0.4  &2& 91.86 & 84.84 & 87.43 \\
    0.5  &3& 91.41 & 88.57 & 88.12\\ 
    0.7  &4& 91.18 & 88.45 &88.19\\
    0.9  &5& 91.58 & 86.64 & 88.18\\
    \bottomrule
  \end{tabular}
  }
  \end{minipage}
\end{table}

\section{Conclusion and limitations}\label{sec: conclusion and limitations}

We introduced DiffoRA, a novel and theoretically-grounded PEFT method that enables module-wise LoRA adoption. Instead of adjusting the interior ranks, DiffoRA employs a selective matrix DAM to identify essential modules, backed by theoretical analysis on convergence and generalization. The DAM is constructed through continuous relaxation and discretization, with a weight-sharing strategy to mitigate discretization discrepancies. Experiments across diverse benchmarks show that DiffoRA consistently outperforms prior methods.
Currently, the sample rate is determined by empirical observations, which differs across models and datasets. This might lead to suboptimal results. In future work, we aim to investigate principled strategies for sample rate selection to improve generalization and adaptability of our method.



{
    \small
\bibliographystyle{IEEEtranN}
\bibliography{IEEEabrv,main}

}

\appendix


\clearpage

\section{Detailed Proofs}
\label{app:proof}

\subsection{Proof of Theorem \ref{theorem:convergence}}
\label{app:sub:th2}

We first introduce the main theorem in \cite{dugradient} and Wely inequality as follows.

\begin{theorem}[Theorem 3.3 in \cite{dugradient}] \label{theorem: du}
If gram matrix $\bm{H}^\infty\succ 0$, $\|\bm{x}_i\|_2=1$, $|y_i|<C$ for some constant $C$ and $i\in[n]$, hidden nodes $m=\Omega\left( \frac{n^6}{\lambda_{\min}(\bm{H}^\infty)^4\delta^3}\right)$, and i.i.d. initialize $\bm{w}_r\sim N(\bm{0},\bm{I})$, $a_r\sim U\{[-1, 1]\}$ for $r\in[m]$, then with probability at least $1-\delta$ over the initialization, the following inequality holds:
\begin{equation}
\begin{aligned}
    \|f(\bm{W}(t),\bm{a},\bm{X})-\bm{y}\|_2^2 \leq \exp{(-\lambda_{\min}(\bm{H}^\infty) t)}\|f(\bm{W}(0),\bm{a},\bm{X})-\bm{y}\|_2^2
\end{aligned}
\end{equation}
where $$\bm{H}^\infty:=\mathbb{E}_{\bm{w}\sim N(\bm{0},\bm{I})}[\bm{x}_i^T\bm{x}_j\mathbb{I}\{\bm{w}^T\bm{x}_i\geq 0,\bm{w}^T\bm{x}_j\geq 0\}].$$
\end{theorem}

The inequality in the above theorem demonstrates that the minimum eigenvalue of the Gram matrix positively affects the training convergence rate of the network.

\begin{lemma}[Weyl inequality \cite{horn2012matrix}]\label{weyl} Let $\bm{A}$, $\bm{B}\in\mathbb{R}^{n\times n}$ be Hermitian matrices, and let the eigenvalues of $\bm{A}$, $\bm{B}$, and $\bm{A} + \bm{B}$ be $\{\lambda_i(\bm{A})\}_{i=1}^n$, $\{\lambda_i(\bm{B})\}_{i=1}^n$ and $\{\lambda_i(\bm{A}+\bm{B})\}_{i=1}^n$, respectively. The eigenvalues of each matrix are arranged in ascending order. Then we have

\begin{equation}\label{weyl:1}
    \lambda_i(\bm{A}+\bm{B})\leq \lambda_{i+j}(\bm{A})+\lambda_{n-j}(\bm{B}),\quad j=\{0,1,\ldots,n-i\}
\end{equation}
for each $i\in [n]$, with equality for some pair $i,j$ if and only if there is a nonzero vector $\bm{x}$
such that $\bm{A}\bm{x}=\lambda_{i+j}(\bm{A})\bm{x}$, $\bm{B}\bm{x}=\lambda_{n-j}(\bm{B})\bm{x}$, and $(\bm{A}+\bm{B})\bm{x}=\lambda_{i}(\bm{A}+\bm{B})\bm{x}$. Also,

\begin{equation}\label{weyl:2}
    \lambda_{i-j+1}(\bm{A})+\lambda_j(\bm{B})\leq \lambda_{i}(\bm{A}+\bm{B}),\quad j=\{1,\ldots, i\}
\end{equation}
for each $i \in [n]$, with equality for some pair $i, j$ if and only if there is a nonzero vector $\bm{x}$ such that $\bm{A}\bm{x}=\lambda_{i-j+1}(\bm{A})\bm{x}$, $\bm{B}\bm{x} =\lambda_j (\bm{B})\bm{x}$, and $(\bm{A} + \bm{B})\bm{x} = \lambda_i (\bm{A} + \bm{B})\bm{x}$. 
If $\bm{A}$ and $\bm{B}$ have no common eigenvector, then inequality (\ref{weyl:1}) and (\ref{weyl:2}) are strict inequality. 
\end{lemma}

We first present and proof the following lemma.

\begin{lemma}
If $\bm{x}_i\nparallel \bm{x}_j, \forall i\neq j$, we have $\bm{H}^\infty_{\bm{w}_0}\succ 0$. 
\end{lemma}

\begin{proof}
By the Lemma 3.4 of \cite{dugradient}, there exists $\bm{w}\sim\mathcal{N}(\bm{0},\bm{I})$, such that when $m$ is sufficiently large, $\|\bm{w}-\bm{w}_0\|$ is sufficiently small.
Then according to the proof of Theorem 3.1 of \cite{dugradient}, we get $\lambda_{\min}(\bm{H}_{\bm{w}_0}^\infty) > 0$.
\end{proof}

Then we provide the proof of Theorem \ref{theorem: genera}.

\noindent\textbf{Theorem \ref{theorem: genera}.} Suppose $f$ is an NN with a single hidden layer and ReLU activation function. 
    Assume $\bm{X}\in \mathbb{R}^{d\times n}$, $\mathbf{w}(0)\sim N(\mathbf{0},\mathbf{I})$, hidden nodes $m=\Omega\left(\frac{n^6d^2}{(\lambda_0^{\bm{\Gamma}})^4\delta^3}\right)$, and $\bm{I}^{\bm{\bm{\Gamma} w}}-\bm{I}^{\bm{w}}\succeq 0$, then the following formula holds with probability at least $1-\delta$ over the initialization
    \begin{equation}
    \begin{aligned}
        &\|f(\mathbf{W}(t),\mathbf{a},\mathbf{X};\bm{\Gamma},\bm{W}_0)-\bm{y}\|_2^2\\
        \leq& \exp(-\lambda_0^{\bm{\Gamma}} t)\|f(\mathbf{W}(0),\mathbf{a},\mathbf{X};\bm{\Gamma},\bm{W}_0)-\bm{y}\|_2^2\\
    \end{aligned}
    \end{equation}
    where $\lambda_0^{\bm{\Gamma}}\geq\lambda_0$. 

\begin{proof}
    We denote $\bm{I}^-:=\bm{I}^{\bm{\bm{\Gamma} w}}-\bm{I}^{\bm{w}}\succeq 0$.
    Then we have the following inequalities:
    \begin{equation}
        \begin{aligned}
            \lambda_{\min}(\bm{H}^\infty_{\bm{\Gamma},\bm{w}_0})&=\lambda_{\min}(\bm{H}^\infty_{\bm{w}_0}+\bm{H}^\infty_{\bm{\Gamma},\bm{w}_0}-\bm{H}_{\bm{w}_0}^\infty)\\
            &\geq \lambda_{\min}(\bm{H}_{\bm{w}_0}^\infty)+\lambda_{\min}(\bm{H}^\infty_{\bm{\Gamma},\bm{w}_0}-\bm{H}_{\bm{w}_0}^\infty)
        \end{aligned}
    \end{equation}
    From the definitions of $\bm{H}^\infty_{\bm{\Gamma},\bm{w}_0}$ and $\bm{H}^\infty_{\bm{w}_0}$ we have

    \begin{equation}
        \bm{H}^\infty_{\bm{\Gamma},\bm{w}_0}-\bm{H}_{\bm{w}_0}^\infty = \bm{X}^T\bm{X}\odot \bm{I}^-
    \end{equation}

    Since $\bm{X}^T\bm{X}$ and $\bm{I}^-$ are both positive definite/semi-positive definite matrices, their Hadamard product is also a positive definite/semi-positive definite matrix \cite{matrixmagnus}, \ie, $\bm{H}^\infty_{\bm{\Gamma},\bm{w}_0}-\bm{H}^\infty_{\bm{w}_0}\succeq 0$.
    Therefore, we have 

        \begin{equation}
        \lambda_{\min}(\bm{H}_{\bm{\Gamma},\bm{w}_0}^\infty)\geq \lambda_{\min}(\bm{H}_{\bm{w}_0}^\infty)
    \end{equation}

    Finally, according to Theorem 1, we have 

    \begin{equation}
        \begin{aligned}
            &\|f(\mathbf{W}(t),\mathbf{a},\mathbf{X};\bm{\Gamma},\bm{W}_0)-\bm{y}\|_2^2\\
        \leq& \exp(-\lambda_0^{\bm{\Gamma}} t)\|f(\mathbf{W}(0),\mathbf{a},\mathbf{X};\bm{\Gamma},\bm{W}_0)-\bm{y}\|_2^2\\
        \end{aligned}
    \end{equation}
    where $\lambda_0^{\bm{\Gamma}} \geq \lambda_0$.
\end{proof}


\begin{table}
  \centering
    \caption{Training settings for GLUE benchmarks.}
  \label{tab: training settings}
  \begin{tabular}{c|cccccccc}
    \toprule
    Datasets & learning rate & batch size & epochs  & r & K &$\alpha$ & LoRA drop & warm-up \\
    \midrule
    MNLI & 3e-4 & 64 & 7  & 4 & 3 & 16 & 0.25 & 4000  \\
    RTE & 5e-4 & 16 & 50  & 4 & 3 & 16 &0.1 & 200\\
    QNLI& 6e-4 & 64 & 5  & 6 & 3 & 16 & 0.35&3000\\
    MRPC & 5e-4 & 32 & 30  & 4 & 3 & 16 & 0 & 200 \\
    QQP & 4e-4 & 32 & 20  & 4 & 3 & 16 & 0 &2000\\
    SST-2 & 1e-4 & 32 & 5& 4 & 3 & 16 & 0 & 3000 \\
    CoLA & 2e-4 & 8 & 20 & 4 & 3 & 16 & 0 & 200 \\
    STS-B & 5e-4 & 16 & 20  & 4 & 2 & 16 & 0.2 & 300 \\
    \bottomrule
  \end{tabular}

\end{table}


\begin{table}
  \centering
    \caption{Training settings for SQuAD benchmarks.}
  \label{tab: training details for qa}
  \begin{tabular}{@{}c|cccc|cccc@{}}
    \toprule
    Settings & \multicolumn{4}{c|}{SQuADv1.1} & \multicolumn{4}{c}{SQuADv2.0}\\
     \midrule
    \#Params &0.08\% &0.16\%& 0.32\% & 0.65\% &0.08\% &0.16\%& 0.32\% & 0.65\%\\
    r & 1 & 2 & 5 & 10 & 2 & 4 & 8 & 15 \\
    \midrule
    train epochs&\multicolumn{4}{c|}{3} & \multicolumn{4}{c}{3}\\
    learning rate&\multicolumn{4}{c|}{5e-4} & \multicolumn{4}{c}{7e-4}\\
    warm-up&\multicolumn{4}{c|}{2000} & \multicolumn{4}{c}{4000}\\
    share r &\multicolumn{4}{c|}{1} & \multicolumn{4}{c}{2}\\

    K & \multicolumn{4}{c|}{3} & \multicolumn{4}{c}{3}\\
    $\alpha$ & \multicolumn{4}{c|}{16} & \multicolumn{4}{c}{16} \\
    LoRA drop & \multicolumn{4}{c|}{16} & \multicolumn{4}{c}{16} \\
    \bottomrule
  \end{tabular}

\end{table}

\subsection{Proof of Theorem \ref{theorem: genera}}
\label{app:sub:th3}

\noindent\textbf{Theorem \ref{theorem: genera}.}
    For an over-parameterized neural network with the loss on the testing set as $\mathcal{L}(\bm{W},\bm{a};\bm{\Gamma},\bm{W}_0)$. Let $\bm{y}=(y_1, ..., y_N)^T$, and $\eta = \kappa C_1\sqrt{\bm{y}^T(\bm{H}_{\bm{\Gamma},\bm{w}_0}^\infty)^{-1}\bm{y}}/(m\sqrt N)$ for some small enough absolute constant $\kappa$, where $\eta$ denotes the step of SGD. Under the assumption of Theorem \ref{theorem:convergence}, for any $\delta\in(0,e^{-1}]$, there exists $m^\ast(\delta,N,\lambda_0^{\bm{\Gamma}})$, such that if $m\geq m^*$, then with probability at least $1-\delta$, we have
    \begin{equation}
    \begin{aligned}
        \mathbb{E}[\mathcal{L}(\bm{W},\bm{a};\bm{\Gamma},\bm{W}_0)]\leq&  \mathcal{O}(C'\sqrt{\frac{\bm{y}^T \bm{y}}{\lambda_0^{\bm{\Gamma}} N}})+ \mathcal{O}(\sqrt{\frac{\log(1/\delta)}{N}})\\
    \end{aligned}
    \end{equation}
    where $\lambda_0^{\bm{\Gamma}}\geq\lambda_0$, $C, C'$, and $\delta$ are constants. 
\begin{proof}
    According to the courant minimax principle \cite{golub2013matrix}, D.2 in \cite{zhu2022generalization}, we get 
\begin{equation*}
\bm{y}^T(\mathbf{H}_{\bm{\Gamma},\bm{w}_0}^\infty)^{-1}\bm{y}\leq\frac{\bm{y}^T\bm{y}}{\lambda_{\min}(\mathbf{H}_{\bm{\Gamma},\bm{w}_0}^\infty)}.
\end{equation*}

Thus, we have 

\begin{align*}
    &\mathbb{E}[\mathcal{L}(\bm{W},\bm{a};\bm{\Gamma},\bm{W}_0)]\\
    &\leq  O\left(c\cdot \sqrt{\frac{\bm{y}^T(\mathbf{H}_{\bm{\Gamma},\bm{w}_0}^\infty)^{-1}\bm{y}}{N}}\right)+O\left(\sqrt{\frac{\log(1/\delta)}{N}}\right)\\
    &\leq O\left(c\cdot \sqrt{\frac{\bm{y}^T\bm{y}}{N\lambda^{\bm{\Gamma}}_0}}\right)+O\left(\sqrt{\frac{\log(1/\delta)}{N}}\right)
\end{align*}
\end{proof}


\begin{table}
  \centering
    \caption{The results of the fine-tuned RoBerta-base model on the GLUE dataset are presented, with the best results highlighted in bold and the second-best results underlined.}
  \label{tab: gluerobert}
  \resizebox{\textwidth}{!}{
  \begin{tabular}{@{}l|c|ccccccccc@{}}
    \toprule
    \multirow{2}{*}{Method} & \multirow{2}{*}{\#Params} & MNLI & SST-2 & CoLA & QQP & QNLI & RTE & MRPC & STS-B & All \\
     & & Acc & Acc & Mcc & Acc & Acc & Acc & Acc & Corr & Avg.\\
    \midrule
    Full FT & 124.65M & 87.68 & 94.73 & 60.26 & 90.75 & 92.58 & 78.63 & 88.33 & 90.31 &85.41\\
    \midrule
    BitFit & 0.1M & 85.50 & 94.38 & 61.16 & 88.08 & 90.99 & 79.57 & 89.07 & 90.55 & 84.91 \\
    \midrule
    HAdapter &1.20M& 86.53& 93.73& 62.62 & \underline{90.83} & \underline{92.82} & 80.43 & \underline{89.90} & 90.16 & 85.88 \\
    PAdapter& 1.19M & 86.75& 93.83 & \underline{63.87} & 90.53 & 92.61 & 80.51 & 89.51& 90.65 & 86.03\\
    LoRA & 1.33M & 87.11 & 93.71 & 63.54 & 90.44 & 92.76 & 80.65 & \underline{89.90} & 90.91& 86.13\\
    AdaLoRA & 1.27M & \textbf{87.89} & \underline{95.11}& 63.23 & 90.48 & \textbf{92.84} & 81.23 & 89.02 & \underline{91.22} & \underline{86.38} \\
    FLoRA & 1.33M & 87.43 & 94.27 & 63.31 & 90.38 & 92.75 & \underline{81.59} & \textbf{90.44} & 90.82 & 86.37\\
    DiffoRA& 1.32M &\underline{87.73} &\textbf{95.16} & \textbf{64.95} &\textbf{91.04} & \textbf{92.84} & \textbf{81.98} & 89.44& \textbf{91.35}&  \textbf{86.81}\\
    \bottomrule
  \end{tabular}
  }

\end{table}

\section{Baseline details}\label{app:baseline}
We use the following methods as baselines. 
(i) Full FT uses all parameters for fine-tuning;
(ii) BitFit \cite{zaken2022bitfit} is a sparse-fine-tuning method for pre-trained models that updates only a small subset of the bias terms;
(iii) Houlsby adapter \cite{houlsby2019parameter} adds a few trainable modules inserted between layers of a pre-trained model, allowing for task-specific tuning without altering the entire model; 
(iv) Pfeiffer adapter \cite{pfeiffer2021adapterfusion} combines multiple task-specific adapters by linearly blending their outputs;
(v) LoRA \cite{hu2022lora} reduces the number of trainable parameters by applying low-rank matrix decomposition to weight updates in pre-trained models;
(vi) AdaLoRA \cite{zhang2023adaptive} adapts LoRA by dynamically adjusting the rank of low-rank updates during training, optimizing parameter efficiency while maintaining model performance across various tasks. 
(vii) LoRA+ \cite{hayou2024lora+} employs different learning rates to update the low-rank matrices.

\section{Dataset details}

\noindent\textbf{GLUE \cite{wang2018glue}} The GLUE Benchmark is a comprehensive collection of natural language understanding tasks, designed to evaluate the performance of models across various NLP applications. It includes:
\begin{itemize}
    \item MNLI: Multinomial Natural Language Inference (inference task), including 393k training data and 20k test data.
   \item SST-2: Stanford Sentiment Treebank (sentiment analysis task), including 67k
   training data and 1.8k test data.
   \item MRPC: Microsoft Research Paraphrase Corpus (paraphrase detection task), including 3.7k training data and 1.7k test data.
   \item CoLA: Corpus of Linguistic Acceptability (linguistic acceptability task), including 8.5k training data and 1k test data.
    \item QNLI: Question Natural Language Inference (inference task), including 108k training data and 5.7k test data.
    \item QQP: Quora Question Pairs (question-answering task), including 364k training data and 391k test data.
    \item RTE: Recognizing Textual Entailment (inference task), including 7k training data and 1.4k test data.
    \item STS-B: Semantic Textual Similarity Benchmark (textual similarity), including 7k training data and 1.4k test data.
\end{itemize}

\noindent\textbf{SQuAD}
SQuADv1.1 \cite{rajpurkar-etal-2016-squad} is a dataset consisting of approximately 100k question-answer pairs based on a collection of Wikipedia articles. The task is to extract exact spans of text from the articles as answers to the given questions.
SQuADv2.0 \cite{rajpurkar-etal-2018-know} builds on v1.1 by adding unanswerable questions. It includes about 150k question-answer pairs from over 500 articles, requiring models to both extract answers and identify when no answer is available.

\section{Training details}
\label{app:settings}
DiffoRA is fully implemented in Python. We evaluate the method on a Desktop Core i7-12700F CPU and Tesla V100 GPU.
We summarize the training settings of GLUE and SQuAD in \cref{tab: training settings} and \cref{tab: training details for qa}, respectively.
In order to compare with the baseline method at the same parameter level, we use different $r$ for different datasets.

\section{Algorithm pseudocode}
\label{app:alg}

We present the training algorithm in Algorithm \ref{alg: difflora}.

\begin{algorithm}[H]
\caption{DiffoRA}
\label{alg: difflora}
\begin{algorithmic}[1] 
\REQUIRE Pre-trained model with $L$ layers $\mathcal{M}^L$; Candidate fine-tuning modules $M$, where $|M|=L\times N$, $N$ is the number of candidate modules in each layers; Training dataset and valid dataset $\bm{X}_{train}$ and $\bm{X}_{valid}$; Training epochs and valid epochs, $T$ and $V$; The learning rate $\eta$; Sample rate $\rho$; LoRA rank: $r_l$; Share rank: $r_s$.
\STATE // Stage 1: Continuous Relaxation
\STATE Create the hyperparameters $\bar{\bm{\Gamma}}\in\mathbb{R}^{L\times N}$.
\FOR{$v=1;v<V;v++$}
\STATE Update hyperparameters $\bar{\bm{\Gamma}}$ as 
$
    \bar{\bm{\Gamma}}_v=\bar{\bm{\Gamma}}_v- \eta \nabla_{\bar{\bm{\Gamma}}}\mathcal{L}_{valid}(\bm{X}_{valid}; \Delta \bm{W}, \bar{\bm{\Gamma}}_{v-1})
$
\FOR{$t=1;t<T;t++$}
\STATE Update low-rank matrix $\Delta \bm{W}$ as follow:
$$
    \Delta \bm{W} = \Delta \bm{W} - \eta \nabla_{\Delta \bm{W}} \mathcal{L}_{train}(\bm{X}_{train};\Delta \bm{W}, \bar{\bm{\Gamma}}_v)
$$
\ENDFOR
\ENDFOR

\STATE // Stage 2: Discretization and Fine-Tuning
\STATE Select the Top-K modules of each layer in $\mathcal{M}^L$ according to the $\bar{\bm{\Gamma}}$, where $K=\lfloor\rho \cdot N\rfloor$.
\STATE Add a low-rank matrix $\Delta \bm{W}=\bm{B A}$ to the selected module and add the weight sharing matrix $\Delta \bm{W}_s = \bm{B}_s \bm{A}_s $ to the remaining modules, where $\bm{B}^T, \bm{A}\in\mathbb{R}^{d\times r_l}$, and $\bm{B}^T_s, \bm{A}_s\in \mathbb{R}^{d\times r_s}$.
\FOR{$t=1;t<T;t++$}
\STATE Update low-rank matrix $\Delta \bm{W}$ as follow:
$$
    \Delta \bm{W} = \Delta \bm{W} - \eta \nabla_{\Delta \bm{W}} \mathcal{L}_{train}(\bm{X}_{train};\Delta \bm{W})
$$

\ENDFOR
\RETURN Fine-tuned model $\mathcal{M}^L$.
\end{algorithmic}
\end{algorithm}

\section{Results on RoBERTa-base}
\label{app:roberta}
Finally, to further prove the effectiveness of our proposed DiffoRA, we adopt RoBERTa-base as the backbone model which contains 125 million parameters.
We fine-tune the pre-trained model on eight tasks of the GLUE benchmark.
We set the rank of the low-rank decomposition matrices to $4\sim12$ and fine-tuned $20\%\sim 70\%$ of the modules in each layer of the pre-trained model.
The experimental results are summarized in \cref{tab: gluerobert}. 
It can be shown that our DiffoRA achieves the best results in almost all datasets.
For instance, our approach outperforms the state-of-the-art method PAdapter on the CoLA dataset by 0.98\%. Although DiffoRA achieves the second-best result on a few tasks, we achieve the highest average accuracy among all the baselines, i.e., 86.81\% on average.  
The experimental results demonstrate the effectiveness of our method.

\newpage
\section*{NeurIPS Paper Checklist}

\begin{enumerate}

\item {\bf Claims}
    \item[] Question: Do the main claims made in the abstract and introduction accurately reflect the paper's contributions and scope?
    \item[] Answer: \answerYes{} 
    \item[] Justification: Our contributions and scope can be accurately reflected in the abstract and introduction.

\item {\bf Limitations}
    \item[] Question: Does the paper discuss the limitations of the work performed by the authors?
    \item[] Answer: \answerYes{} 
    \item[] Justification: The limitations can be found in \cref{sec: conclusion and limitations}.

\item {\bf Theory assumptions and proofs}
    \item[] Question: For each theoretical result, does the paper provide the full set of assumptions and a complete (and correct) proof?
    \item[] Answer: \answerYes{} 
    \item[] Justification: We provide the full set of assumptions and a complete proof in the \cref{app:proof}.

    \item {\bf Experimental result reproducibility}
    \item[] Question: Does the paper fully disclose all the information needed to reproduce the main experimental results of the paper to the extent that it affects the main claims and/or conclusions of the paper (regardless of whether the code and data are provided or not)?
    \item[] Answer: \answerYes{} 
    \item[] Justification: We provided a README file in our code for the reviewers to check.

\item {\bf Open access to data and code}
    \item[] Question: Does the paper provide open access to the data and code, with sufficient instructions to faithfully reproduce the main experimental results, as described in supplemental material?
    \item[] Answer: \answerYes{} 
    \item[] Justification:  We provided an anonymous URL link for our codes in \cref{subsec: configurations}. All the datasets we used are public.

\item {\bf Experimental setting/details}
    \item[] Question: Does the paper specify all the training and test details (e.g., data splits, hyperparameters, how they were chosen, type of optimizer, etc.) necessary to understand the results?
    \item[] Answer: \answerYes{} 
    \item[] Justification: We have described our experimental settings in \cref{subsec: configurations}. For additional experimental settings on various benchmarks, we provided the details in \cref{app:settings}.

\item {\bf Experiment statistical significance}
    \item[] Question: Does the paper report error bars suitably and correctly defined or other appropriate information about the statistical significance of the experiments?
    \item[] Answer: \answerYes{} 
    \item[] Justification: We have clearly stated the factors of variability that the error bars are capturing. We have run our experiments multiple times with different seeds and provided the mean and standard deviation values in Section 5.

\item {\bf Experiments compute resources}
    \item[] Question: For each experiment, does the paper provide sufficient information on the computer resources (type of compute workers, memory, time of execution) needed to reproduce the experiments?
    \item[] Answer: \answerYes{} 
    \item[] Justification: We have provided the hardware configurations in \cref{app:settings}.
    
\item {\bf Code of ethics}
    \item[] Question: Does the research conducted in the paper conform, in every respect, with the NeurIPS Code of Ethics \url{https://neurips.cc/public/EthicsGuidelines}?
    \item[] Answer: \answerYes{} 
    \item[] Justification: We confirm that we preserve the anonymity and conform to the NeurIPS Codes of Ethics.

\item {\bf Broader impacts}
    \item[] Question: Does the paper discuss both potential positive societal impacts and negative societal impacts of the work performed?
    \item[] Answer: \answerNA{} 
    \item[] Justification: This work is fundamental research in computer science and has no societal impact.
    
\item {\bf Safeguards}
    \item[] Question: Does the paper describe safeguards that have been put in place for responsible release of data or models that have a high risk for misuse (e.g., pretrained language models, image generators, or scraped datasets)?
    \item[] Answer: \answerNA{} 
    \item[] Justification: This paper poses no risks above.

\item {\bf Licenses for existing assets}
    \item[] Question: Are the creators or original owners of assets (e.g., code, data, models), used in the paper, properly credited and are the license and terms of use explicitly mentioned and properly respected?
    \item[] Answer: \answerYes{} 
    \item[] Justification: All the codes and datasets used in this paper are open-sourced. We have properly cited these resources in our paper. 

\item {\bf New assets}
    \item[] Question: Are new assets introduced in the paper well documented and is the documentation provided alongside the assets?
    \item[] Answer: \answerYes{} 
    \item[] Justification: We have provided an anonymous URL link for our codes.

\item {\bf Crowdsourcing and research with human subjects}
    \item[] Question: For crowdsourcing experiments and research with human subjects, does the paper include the full text of instructions given to participants and screenshots, if applicable, as well as details about compensation (if any)? 
    \item[] Answer: \answerNA{} 
    \item[] Justification: This paper does not involve crowdsourcing or research with human subjects.

\item {\bf Institutional review board (IRB) approvals or equivalent for research with human subjects}
    \item[] Question: Does the paper describe potential risks incurred by study participants, whether such risks were disclosed to the subjects, and whether Institutional Review Board (IRB) approvals (or an equivalent approval/review based on the requirements of your country or institution) were obtained?
    \item[] Answer: \answerNA{} 
    \item[] Justification: This paper does not involve crowdsourcing or research with human subjects.

\item {\bf Declaration of LLM usage}
    \item[] Question: Does the paper describe the usage of LLMs if it is an important, original, or non-standard component of the core methods in this research? Note that if the LLM is used only for writing, editing, or formatting purposes and does not impact the core methodology, scientific rigorousness, or originality of the research, declaration is not required.
    \item[] Answer: \answerNA{} 
    \item[] Justification: We utilize LLM only for grammar and formatting purposes.

\end{enumerate}

\end{document}